\documentclass[letterpaper]{article} % DO NOT CHANGE THIS
\usepackage{aaai2026}  % DO NOT CHANGE THIS
\usepackage{times}  % DO NOT CHANGE THIS
\usepackage{helvet}  % DO NOT CHANGE THIS
\usepackage{courier}  % DO NOT CHANGE THIS
\usepackage[hyphens]{url}  % DO NOT CHANGE THIS
\usepackage{graphicx} % DO NOT CHANGE THIS
\usepackage{natbib}  % DO NOT CHANGE THIS AND DO NOT ADD ANY OPTIONS TO IT
\usepackage{caption} % DO NOT CHANGE THIS AND DO NOT ADD ANY OPTIONS TO IT
\usepackage{algorithm}
\usepackage{algorithmic}

\usepackage{newfloat}
\usepackage{listings}
\DeclareCaptionStyle{ruled}{labelfont=normalfont,labelsep=colon,strut=off} % DO NOT CHANGE THIS
\floatstyle{ruled}
\newfloat{listing}{tb}{lst}{}
\floatname{listing}{Listing}

\usepackage{tikz}
\usetikzlibrary{arrows.meta, positioning, fit}
\usepackage{booktabs}
\usepackage{subcaption}
\usepackage{makecell}

\usepackage{listings}

\makeatletter
\newcommand{\zhmkbox}[2]{%
  \expandafter\newsavebox\csname zh@#1\endcsname
  \expandafter\sbox\csname zh@#1\endcsname{%
    \begin{CJK*}{UTF8}{gbsn}\ttfamily\footnotesize #2\end{CJK*}}%
}
\newcommand{\zhuse}[1]{\copy\csname zh@#1\endcsname}
\makeatother

\title{A Human-Centred Approach to Benchmarking LLMs for Parenting Advice}
\author {
    Yunke Zhao\equalcontrib\textsuperscript{\rm 1},
    Isobel Voysey\equalcontrib\textsuperscript{\rm 1},
    Alastair van Heerden\textsuperscript{\rm 2},
    Rob Hughes\textsuperscript{\rm 2},
    Jun Zhao\textsuperscript{\rm 1}
}
\affiliations {
    \textsuperscript{\rm 1}Department of Computer Science, University of Oxford\\
    \textsuperscript{\rm 2}Tandem\\
}

\newcommand{\rob}[1]{}

\begin{document}

\maketitle

\begin{abstract}
People are increasingly using large language models (LLMs) to seek advice, including for parenting. Parenting is a critical and socially sensitive domain. Thus, evaluating advice provided by LLMs requires indicators beyond aggregated information quality benchmarks to consider relational and behavioural elements of the responses. With a multi-dimensional rubric created by parenting experts, this paper evaluates 15 LLMs across 100 parenting scenarios in 2 languages (English and Chinese), using an LLM-as-a-judge method. Results show that aggregate scores can hide rubric item-specific weaknesses, models implicitly encourage different parenting styles, and language influences responses. We highlight the importance of evaluation output auditability and challenges involved in evaluating LLM-generated advice in domains like parenting. Our findings provide important insights for selecting LLMs for direct user engagement and the development of user-facing parenting advice applications. 
\end{abstract}


\section{Introduction}

Parents routinely make complex decisions about how best to care for their children. To support these decisions, parents have long relied on advice from their family, friends, and professionals, and a large proportion seek information and advice through online sources, including social media, parenting forums, and traditional search engines \cite{mertensParentingInformationSocial2024, bakerWhoUsesOnline2017}. More and more, large language models (LLMs) such as ChatGPT are becoming part of this informational ecosystem, with early evidence suggesting that some parents are beginning to use LLMs to seek parenting guidance and support~\cite{sharminAvoidingSocialJudgment2026}\footnote{Further evidence includes popular media depictions of parents using LLMs for advice, such as a recent subplot in \textit{Amandaland}, a British comedy series about parenting~\cite{amandaland}.}.

Existing research evaluating LLMs for parents or in parenting-adjacent domains has largely focused on health information tasks, where outputs can often be assessed against relatively clear standards of factual correctness, safety, or completeness ~\cite{mcfaydenChatGPTArtificialIntelligence2024, bushuvenChatGPTCanYou2023, sezginClinicalAccuracyLarge2023}. However, the quality of parenting advice extends beyond factual accuracy alone. Parenting is a socially and culturally situated practice shaped by values, norms, beliefs about child development, and differing theories of parent-child relationships \cite{linParentingCulturesIdealParent2023, lansfordAnnualResearchReview2022}. In many parenting situations, there may be no single objectively correct answer. Instead, advice quality may be affected by how actionable it is, the tone taken, and how relatable it is to specific cultural contexts or long-term challenges facing a family.

These characteristics create challenges for evaluating LLM-generated parenting advice. Traditional benchmark approaches centred on accuracy or matching human preferences~\cite{beanMeasuringWhatMatters2025} may fail to capture the relational and value-laden nature of parenting support. Evaluating parenting advice therefore requires more \textbf{human-centred approaches} that consider not only whether advice is safe and accurate, but also social qualities of the support provided by the model, what parenting style the model implicitly reinforces, and how advice differs across cultural and linguistic contexts.

In this paper, we present a multilingual evaluation pipeline for assessing LLM-generated parenting advice. The pipeline combines scenario generation and refinement, large-scale response generation, rubric-based evaluation using LLM judges, and parenting-style analysis grounded in established parenting theory. Using 100 parenting scenarios in English and Mandarin (Chinese), we compare responses from 15 LLMs across multiple dimensions of advice quality and parenting style.

Our work addresses the following questions:

\begin{enumerate}
    \item How does rubric-based evaluation reveal different forms of user-facing advice quality across models and languages?
    % \item How do LLMs differ behaviourally in how they position, guide, reassure, and structure parent-facing advice interactions?
    \item How implicit parenting styles are reflected in LLM-generated advice, and how does this vary across models and languages?
\end{enumerate}

Through our work we make three primary contributions:

\begin{itemize}
    \item \textbf{A scalable multilingual pipeline for evaluating LLM-generated parenting advice}: The pipeline encompasses scenario generation, response generation, LLM-based judging, and automated analysis. The pipeline also includes various failure repair strategies, an auditable output trace for each score, and the option to conduct the same evaluation in another language.
    \item \textbf{A human-centered evaluation framework for parenting advice}: The expert-informed rubric moves beyond factual correctness to incorporate relational and behavioural qualities of LLM-generated advice, which is further extended through classification of implicit parenting style.
    \item \textbf{Empirical comparison of LLMs across advice quality and parenting styles}: Our findings show a wide range in advice quality across models and model families and models displayed different capability strengths across the eight evaluation rubrics, confirming a multi-dimensionality of LLMs for supporting user-facing scenarios. Furthermore,  when responding in English, models lean more authoritative in parenting style than they do when responding in Chinese, where they lean more authoritarian.
\end{itemize}

% We find substantial variation across models in \todo{advice quality}, including differences in \todo{key rubric measures}. We additionally observe cross-language variation in \todo{advice quality and parenting style}, highlighting the importance of multilingual assessment in socially situated domains.

\section{Background}

\subsection{LLMs and Parenting}

With the global rise of nuclear families, many parents go online to seek information and social support for parenting~\cite{mertensParentingInformationSocial2024, bakerWhoUsesOnline2017, dworkinLiteratureReviewParents2013}, marking a shift from previous generations that predominately relied on families, close friends, and childcare professionals (e.g., teachers, paediatricians).
Developments in LLMs have introduced a new opportunities for parents, especially those from in Western countries, who are in dire need of tools that provide information, advice, and social support~\cite{sharminAvoidingSocialJudgment2026}, necessitating the evaluation of these systems.

Research has not yet established the real-world prevalence of parents using LLMs, but patterns of parents' existing online behaviour combined with high usage among the general public makes LLMs for parenting support worth study.
For example, \citet{yunOnlineHealthInformation2025} found that 21\% of adult survey participants were using LLMs to seek health information, motivating researchers to evaluate LLMs as a tool for parents seeking health information about their child~\cite{mcfaydenChatGPTArtificialIntelligence2024, bushuvenChatGPTCanYou2023, sezginClinicalAccuracyLarge2023, leslie-millerCriticalNeedExpert2024}. 
All of these works indicate the potential for LLMs in these situations, but the authors note shortcomings, like missing or inaccurate references~\cite{sezginClinicalAccuracyLarge2023,mcfaydenChatGPTArtificialIntelligence2024} or the failure to provide appropriate, actionable recommendations~\cite{bushuvenChatGPTCanYou2023, mcfaydenChatGPTArtificialIntelligence2024}.

ChatGPT has been the primary LLM under investigation in these cases. Researchers have looked into the effect of different base models, e.g., GPT-3.5 vs GPT-4~\cite{kimTestingCapabilityGenerative2025, bushuvenChatGPTCanYou2023}, and different data sources, e.g, training data only vs access to internet vs access to parenting-specific content repository~\cite{kimTestingCapabilityGenerative2025} on parenting advice quality, but no work so far has compared a broad set of models from several different families for parenting advice.
To address this gap, we present the first comparison of LLM-generated advice for general-purpose parenting advice from multiple model families.

\subsection{Human-Centered Evaluation of LLMs}

Evaluating LLMs remains a challenge. Traditional approaches have relied on automated benchmarks capturing performance on a narrowly specified task, but this falls short in tasks where good performance is highly subjective~\cite{xiaoHumanCenteredEvaluationAuditing2024}. Researchers have therefore attempted to assess how aligned AI is with human values and preferences, but this presents its own challenges.
Many forms of human-in-the-loop evaluation are time-consuming and resource-intensive, limiting the extent of evaluation. This is reflected in the literature on LLMs in the parenting domain, where evaluations have typically been based on only a small number of questions or vignettes, ranging from as few as 3~\cite{leslie-millerCriticalNeedExpert2024} to 22~\cite{bushuvenChatGPTCanYou2023}.
When it comes to optimising for a set of preferences, differences may be substantial across groups, leading researchers to prompt careful consideration of whose preferences are optimised for in what tasks~\cite{kirkPRISMAlignmentDataset2024a}.
% 5 \cite{kimTestingCapabilityGenerative2025}, 13 \citet{mcfaydenChatGPTArtificialIntelligence2024}, 22 \citet{bushuvenChatGPTCanYou2023}, 3 \citet{leslie-millerCriticalNeedExpert2024}, 14 \citet{sezginClinicalAccuracyLarge2023}

Parenting advice is often subjective, making it challenge to evaluate.
Previous work on LLMs for parents seeking information was largely restricted to the medical domain, where experts largely concur on appropriate and inappropriate courses of action, meaning these works focused mainly on accuracy of information. 
\citet{bushuvenChatGPTCanYou2023} analysed LLM responses to vignettes about paediatric emergencies, coding the diagnosis, advice to call medical professionals, and advice given to first aiders as correct or incorrect.
\citet{mcfaydenChatGPTArtificialIntelligence2024} scored ChatGPT responses to questions about autism on three scales, correctness, clarity, and conciseness. However, providing a set of `correct' answers to compare an LLM response to is not a viable option for more general parenting questions, like suitable discipline practices, particularly given the aforementioned individual and cultural variations in parenting.

Furthermore, the finding by \citet{sharminAvoidingSocialJudgment2026} that some mothers are turning to LLMs for advice to avoid social judgement suggests that relational dimensions of advice, like empathy, are particularly important.
These have largely been neglected in existing work in the parenting domain.
\citet{mcfaydenChatGPTArtificialIntelligence2024} considered how understandable and actionable advice was, as well as whether the language used was more medical or neurodiversity-affirming.
Response characteristics like this may affect parents' perceptions of LLMs; \citet{leslie-millerCriticalNeedExpert2024} found no difference between the trustworthiness and perceived expertise of responses from ChatGPT and paediatricians when parents were blind to the source.

Evaluating parenting advice \textbf{requires consideration of subjective and relational dimensions} that extend beyond simple correctness. LLM-as-a-judge approaches enable scalable assessment of such complex responses, but prior work has identified potential biases, including preferences for longer answers and models in the same family~\cite{stureborgLargeLanguageModels2024, kooBenchmarkingCognitiveBiases2024}. We therefore develop a pipeline that combines multiple LLM judges with an expert-informed multidimensional rubric designed to assess both informational quality and the relational characteristics of advice.

\subsection{Culturally-Sensitive Parenting Advice}

Evaluating LLMs for parenting requires considering not only the relational aspects of the interaction between the LLM and the parent, but also the advised interaction between the parent and the child.
% \todo{Considering the relational dimension between parent and child, \citet{kimTestingCapabilityGenerative2025} coded responses for `connection'---responses that reflected relational complexity or the interelatedness of the child and parent scored higher.}
Parenting is a socially and culturally situated practice \cite{linParentingCulturesIdealParent2023,lansfordAnnualResearchReview2022}, which means that parenting advice tends to go beyond purely informational content to promote certain patterns of behaviour, attitudes, and approaches---normative guidance about how parents should react and relate to their children---commonly referred to as parenting styles. 
What is considered by the majority to be an appropriate or ideal parenting style varies across cultures~\cite{linParentingCulturesIdealParent2023, grusecPerspectivesParentDiscipline2017} and within them, as an individual's parenting decisions are also shaped by their personal experiences~\cite{rutiglianoUnderstandingParentsSelfAwareness2023}.
Furthermore, research suggests the parenting style associated with positive child outcomes is also dependent on culture; authoritarian parenting may result in negative outcomes for children in individualistic societies where autonomy is prioritised but work well in collectivist societies where group needs are paramount~\cite{rudyCorrelatesAuthoritarianParenting2001}.
As such, professionals advocate for understanding and integrating culture in support for parents~\cite{harknessWhyUnderstandingCulture2020,schillingCulturalAdaptationGroup2021, baumannCulturalAdaptationImplementation2015}.

Existing work indicates that generic LLMs are poor at responding in culturally sensitive ways, i.e., providing advice that considers the user's beliefs and attitudes, which are shaped by the values, norms, and assumptions prevalent in their community~\cite{aldaweeshItHasntLived2026}, largely representing WEIRD contexts~\cite{qiu2025evaluating}.
Previous evaluations of LLMs for parenting advice have not brought in culture or context as a relevant factor, usually considering a small set of scenarios with no acknowledgement of or variation in context (e.g., ``How do I set boundaries and discipline my child?''~\cite{kimTestingCapabilityGenerative2025}).
Similarly, LLM performance and response style can vary across languages~\cite{chevi2025individual}, which has not yet been analysed in the parenting domain.

In this work, we extend the existing evaluations through the development of a pipeline that evaluates the LLM across a set of context-specified scenarios and classifies parenting style according to Baumrind, Maccoby, and Martin's responsiveness-demandingness theory \cite{baumrindCurrentPatternsParental1971, maccobySocializationContextFamily1983} in order to indicate models' normative stance on how parents should react to and relate to children. This evaluation is carried out in English and Chinese to further investigate cross-language variation in advice quality and characteristics.

\section{Methods}

\subsection{Overview of Evaluation Pipeline}

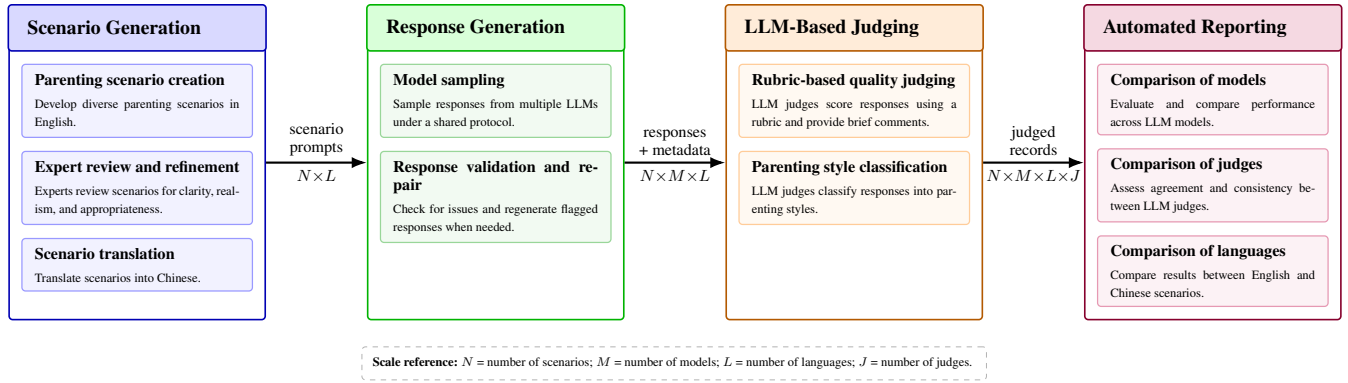
\begin{figure*}[t]
\centering
\resizebox{\linewidth}{!}{
\begin{tikzpicture}[
    font=\small,
    >=Latex,
    stage/.style={
        rounded corners=2pt,
        line width=0.9pt,
        inner sep=0pt,
        fill=white,
        draw=#1!70!black,
    },
    subbox/.style={
        rounded corners=2pt,
        line width=0.7pt,
        inner sep=5pt,
        fill=#1!5,
        draw=#1!40,
    },
    flow/.style={
        -Latex,
        very thick
    }
]

% ---------------------------------------------------------
% Dimensions
% ---------------------------------------------------------

\def\stageW{5cm}
\def\stageH{6.13cm}
\def\boxW{4.5cm}
\def\minipageW{4cm}
\def\boxGap{0.25cm}

% ---------------------------------------------------------
% STAGE 1
% ---------------------------------------------------------

\node[
    stage=blue,
    minimum width=\stageW,
    minimum height=\stageH,
    anchor=north west
] (A) at (0,0) {};

% Header
% \fill[blue!18]
%     (A.north west) rectangle ([yshift=-0.9cm]A.north east);

% \draw[blue!70!black]
%     ([yshift=-0.9cm]A.north west) --
%     ([yshift=-0.9cm]A.north east);

\filldraw[draw=blue!70!black, line width = 0.9pt, inner sep=0pt, fill=blue!15, rounded corners=2pt]
    (A.north west) rectangle ([yshift=-0.9cm]A.north east);

\node[
    anchor=west,
    font=\bfseries\large
] at ([xshift=0.25cm,yshift=-0.45cm]A.north west)
{Scenario Generation};

% Subboxes
\node[
    subbox=blue,
    anchor=north,
    minimum width=\boxW
] (A1) at ([yshift=-\boxGap-0.9cm]A.north)
{
\begin{minipage}{\minipageW}
% \centering
\textbf{Parenting scenario creation}

\vspace{1mm}

\scriptsize
Develop diverse parenting scenarios in English.
\end{minipage}
};

\node[
    subbox=blue,
    anchor=north,
    minimum width=\boxW
] (A2) at ([yshift=-\boxGap]A1.south)
{
\begin{minipage}{\minipageW}
% \centering
\textbf{Expert review and refinement}

\vspace{1mm}

\scriptsize
Experts review scenarios for clarity, realism, and appropriateness.
\end{minipage}
};

\node[
    subbox=blue,
    anchor=north,
    minimum width=\boxW
] (A3) at ([yshift=-\boxGap]A2.south)
{
\begin{minipage}{\minipageW}
% \centering
\textbf{Scenario translation}

\vspace{1mm}

\scriptsize
Translate scenarios into Chinese.
\end{minipage}
};

% ---------------------------------------------------------
% STAGE 2
% ---------------------------------------------------------

\node[
    stage=green,
    minimum width=\stageW,
    minimum height=\stageH,
    anchor=north west
] (B) at (7,0) {};

\filldraw[draw=green!70!black, line width = 0.9pt, inner sep=0pt, fill=green!15, rounded corners=2pt]
    (B.north west) rectangle ([yshift=-0.9cm]B.north east);

% \draw[green!60!black]
%     ([yshift=-0.9cm]B.north west) --
%     ([yshift=-0.9cm]B.north east);

\node[
    anchor=west,
    font=\bfseries\large
] at ([xshift=0.25cm,yshift=-0.45cm]B.north west)
{Response Generation};

\node[
    subbox=green!60!black,
    anchor=north,
    minimum width=\boxW
] (B1) at ([yshift=-\boxGap-0.9cm]B.north)
{
\begin{minipage}{\minipageW}
% \centering
\textbf{Model sampling}

\vspace{1mm}

\scriptsize
Sample responses from multiple LLMs under a shared protocol.
\end{minipage}
};

\node[
    subbox=green!60!black,
    anchor=north,
    minimum width=\boxW
] (B2) at ([yshift=-\boxGap]B1.south)
{
\begin{minipage}{\minipageW}
% \centering
\textbf{Response validation and repair}

\vspace{1mm}

\scriptsize
Check for issues and regenerate flagged responses when needed.
\end{minipage}
};

% ---------------------------------------------------------
% STAGE 3
% ---------------------------------------------------------

\node[
    stage=orange,
    minimum width=\stageW,
    minimum height=\stageH,
    anchor=north west
] (C) at (14,0) {};

\filldraw[draw=orange!70!black, line width = 0.9pt, inner sep=0pt, fill=orange!15, rounded corners=2pt]
    (C.north west) rectangle ([yshift=-0.9cm]C.north east);

\node[
    anchor=west,
    font=\bfseries\large
] at ([xshift=0.25cm,yshift=-0.45cm]C.north west)
{LLM-Based Judging};

\node[
    subbox=orange,
    anchor=north,
    minimum width=\boxW
] (C1) at ([yshift=-\boxGap-0.9cm]C.north)
{
\begin{minipage}{\minipageW}
% \centering
\textbf{Rubric-based quality judging}

\vspace{1mm}

\scriptsize
LLM judges score responses using a rubric and provide brief comments.
\end{minipage}
};

\node[
    subbox=orange,
    anchor=north,
    minimum width=\boxW
] (C2) at ([yshift=-\boxGap]C1.south)
{
\begin{minipage}{\minipageW}
% \centering
\textbf{Parenting style classification}

\vspace{1mm}

\scriptsize
LLM judges classify responses into parenting styles.
\end{minipage}
};

% ---------------------------------------------------------
% STAGE 4
% ---------------------------------------------------------

\node[
    stage=purple,
    minimum width=\stageW,
    minimum height=\stageH,
    anchor=north west
] (D) at (21,0) {};

\filldraw[draw=purple!70!black, line width = 0.9pt, inner sep=0pt, fill=purple!15, rounded corners=2pt]
    (D.north west) rectangle ([yshift=-0.9cm]D.north east);

\node[
    anchor=west,
    font=\bfseries\large
] at ([xshift=0.25cm,yshift=-0.45cm]D.north west)
{Automated Reporting};

\node[
    subbox=purple,
    anchor=north,
    minimum width=\boxW
] (D1) at ([yshift=-\boxGap-0.9cm]D.north)
{
\begin{minipage}{\minipageW}
% \centering
\textbf{Comparison of models}

\vspace{1mm}

\scriptsize
Evaluate and compare performance across LLM models.
\end{minipage}
};

\node[
    subbox=purple,
    anchor=north,
    minimum width=\boxW
] (D2) at ([yshift=-\boxGap]D1.south)
{
\begin{minipage}{\minipageW}
% \centering
\textbf{Comparison of judges}

\vspace{1mm}

\scriptsize
Assess agreement and consistency between LLM judges.
\end{minipage}
};

\node[
    subbox=purple,
    anchor=north,
    minimum width=\boxW
] (D3) at ([yshift=-\boxGap]D2.south)
{
\begin{minipage}{\minipageW}
% \centering
\textbf{Comparison of languages}

\vspace{1mm}

\scriptsize
Compare results between English and Chinese scenarios.
\end{minipage}
};

% ---------------------------------------------------------
% Arrows
% ---------------------------------------------------------

\draw[flow]
(A.east) --
node[midway,above,font=\footnotesize,align=center]
{scenario\\prompts}
node[midway,below,font=\footnotesize,align=center]
{$N{\times}L$}
(B.west);

\draw[flow]
(B.east) --
node[midway,above,font=\footnotesize,align=center]
{responses\\+ metadata}
node[midway,below,font=\footnotesize,align=center]
{$N{\times}M{\times}L$}
(C.west);

\draw[flow]
(C.east) --
node[midway,above,font=\footnotesize,align=center]
{judged\\records}
node[midway,below,font=\footnotesize,align=center]
{$N{\times}M{\times}L{\times}J$}
(D.west);

% ---------------------------------------------------------
% Scale reference
% ---------------------------------------------------------

\node[
    draw=gray!60,
    dashed,
    rounded corners=2pt,
    inner sep=6pt,
    font=\scriptsize,
    align=center
] at (13,-7)
{
\textbf{Scale reference:}
$N$ = number of scenarios;
$M$ = number of models;
$L$ = number of languages;
$J$ = number of judges.
};

\end{tikzpicture}
}
\caption{Simplified evaluation workflow}
\label{fig:simplified-workflow}

\end{figure*}

We developed a multi-stage evaluation pipeline to assess the quality and characteristics of LLM parenting advice across multiple models and languages\footnote{\url{https://github.com/UndeadCZ13/parenting-advice-llm-evals}}. The pipeline consisted of four stages, which are shown in Figure~\ref{fig:simplified-workflow}:
\begin{enumerate}
    \item Expert-guided parenting scenario generation
    \item Response generation
    \item LLM-based judging
    \item Automated quantitative analysis and reporting of benchmarking results
\end{enumerate}

The evaluation was conducted over 15 models using 100 parenting scenarios presented in both English and Chinese. Responses generated by LLMs were assessed using rubric-based quality scoring and parenting-style classification by two different LLM judges.

% =========================================================
% Simplified full evaluation workflow
% =========================================================
%
% Required packages:
%
% \usepackage[a4paper,margin=2cm]{geometry}
% \usepackage{tikz}
% \usepackage{xcolor}
% \usetikzlibrary{arrows.meta,positioning}
%
% =========================================================

% cannot use input in final version

% \begin{figure*}
%     \centering
%     \includegraphics[width=\linewidth]{flowchartinspo.png}
%     \caption{Uneditable, but more like what I want}
%     \label{fig:placeholder}
% \end{figure*}

\subsection{Scenario Generation}

\subsubsection{Parenting Scenario Creation}

An initial set of 100 parenting scenarios was generated using ChatGPT.
\rob{which model?}
The scenarios were designed to represent a broad range of common parenting situations encountered across developmental stages and family contexts.
The generated scenarios covered multiple topics, including health, nutrition, safety, development, discipline, education, sleep, parent wellbeing, social-emotional support, special needs, and low-resource contexts.
Each scenario consisted of a single-turn prompt describing a parenting situation and requesting advice. No follow-up clarification or conversational context was provided, allowing all models to respond under identical single-shot conditions.

\rob{Further details on model used, generation parameters, prompting strategy, distribution across categories, etc}

\subsubsection{Expert Review and Refinement}

The initial scenarios were manually reviewed and refined by an expert group with domain expertise in parenting and child development. The review process aimed to improve realism, diversity, and clarity.
\rob{more details in this section, number of experts, domain expertise, purpose of process}

\subsubsection{Scenario Translation}

All finalised scenarios were translated from English into Chinese to enable cross-language comparison. This translation was carried out using GPT-5 under prompt constraints. The prompt gave explicit instructions to preserve semantic content, retain the original level of risk and tone, avoid adding or removing key facts, and avoid introducing extra cultural assumptions (see Appendix).
A subset of scenarios were reviewed by an author fluent in both English and Chinese to verify the accuracy of the translation.

\subsection{Response Generation}

\subsubsection{Model Sampling}

Parenting advice responses were generated using 15 LLMs across 100 scenarios in 2 languages (English and Chinese).
The models evaluated spanned \textbf{several model families and deployment types}, including frontier systems, compact variants, and smaller open or service-based models.

The set included four GPT-family models (GPT-5.2, GPT-5 Nano, GPT-OSS 20B, and GPT-4o Mini), two DeepSeek models (DeepSeek V3.1 and DeepSeek R1), two Qwen models (Qwen3 32B and Qwen3 8B), two Llama models (Llama 3.3 70B and Llama 3.1 8B), two GLM models (GLM-4.6 and GLM-4 9B), and three additional models: Kimi K2 Thinking, MiniMax M2, and Ministral 3 14B\footnote{Claude models were not included because access was not available through our institution at the time of evaluation, which we acknowledge as a limitation of the work. Extending the comparison to Claude models is an important avenue for future work.}.
% \todo{why not claude, one may ask}

Each model received the full scenario text, including the context, tags, and parent question. The \textit{context} gives relevant information about the child, family, or setting. The \textit{question} defines the advice-seeking task. The \textit{tags} locate the scenario within broader parenting topics. Prompts were language-aligned: English scenarios elicited English responses, and Chinese scenarios elicited Chinese responses. The task was framed as giving practical advice to a parent in the described situation.

The prompt used a final-answer-only design. Models were instructed to provide the parent-facing answer without intermediate reasoning or wrapper text. This reduces format variation across model families and makes the judged response unit more consistent.

\subsubsection{Response Validation and Repair}

To ensure reliable assessment, we implemented a response validation, repair, and cleaning stage to identify incomplete, malformed, or failed generations that could distort later scores for reasons unrelated to advice quality.

Problematic responses were automatically flagged using generation metadata and rule-based heuristics, including API errors, truncation indicators, missing terminal punctuation, cut-off endings, and unusually short final fragments.

Flagged cases were then manually reviewed before rerun, since simple surface rules can over-flag complete responses, especially across languages and punctuation systems. 

Confirmed defective response were regenerated for a maximum of three attempts. Repaired responses were merged back into the canonical answer files, with repair manifests and rerun logs recording which rows were changed. This preserved auditability while reducing the risk that execution failures contaminated model comparison.

A further cleaning step was applied after generation because different model families often produce different output shapes. Reasoning-like segments, wrapper formats such as \verb|<think>| blocks, and explicit final-answer markers were stripped from the raw output to remove this as a confounding factor from judging. The original model output was stored, alongside the cleaned output and further metadata recording if and how many characters were removed.

To further enhance auditability, generation metadata was recorded alongside each response. This included finish reason, retry use, suspected truncation flags, and API-related errors. These signals helped distinguish complete but weak answers from outputs affected by backend failure, empty generation, or length-limited stopping.

\subsection{LLM-Based Judging}

Generated parenting advice was evaluated using an LLM-as-a-judge framework employing two independent judge models: GPT-5.2 as the primary judge and DeepSeek V3.1 as the secondary judge. The secondary judge provided a robustness check from a different model family.
During the judge process, each LLM-generated response was \textbf{assessed across eight predefined quality rubrics}, measuring dimensions of parenting advice quality, and then further were \textbf{classified by parenting style}. As a result, the judging stage produced quantitative rubric scores, parenting style classifications, and qualitative annotations for scoring decisions.

\subsubsection{Rubric-Based Judging}

The pipeline used LLM-as-a-judge as the main scoring method because parenting advice is open-ended and cannot be evaluated through a single reference answer. The judging stage used eight rubrics for advice quality inspired by HealthBench~\cite{aroraHealthBenchEvaluatingLarge2025} and informed by parenting experts. They include the follows:
\begin{enumerate}
    \item Accuracy: factual correctness and evidence basis
    \item Safety: harm avoidance and risk awareness
    \item Helpfulness: actionable, practical steps
    \item Empathy: supportive, non-judgmental tone
    \item Completeness: covers key aspects without major omissions
    \item Bias avoidance: avoids stereotypes and harmful assumptions
    \item Limitation awareness: recognises limits and refers to professionals when appropriate
    \item Communication: presents clear structure and asks clarifying questions when needed
\end{enumerate}

These rubrics allow response quality to be analysed across multiple dimensions. They make it possible to distinguish, for example, between advice that is practical but weak in safety and advice that is safe but too vague to be useful. 

The judge prompt presented the scenario, the model response, and the rubric definitions. The judge assigned a score from 0 to 100 for each rubric and provided one short comment explaining the judgement. The output was constrained to a fixed JSON schema so that scores and comments could be parsed and merged consistently. 
The primary LLM judge evaluated each response three times, and the final score for each rubric item was computed as the mean across repeats. The overall score was the mean across the eight rubric items.
This process was repeated for the secondary judge.

%The short LLM judge comment was retained as qualitative support for later interpretation. It helped explain scenario-level score differences, but it was treated as model-generated evidence rather than human expert annotation. 

\subsubsection{Parenting Style Classification}

Rubric scores describe response quality, but they do not fully capture the nature of the advice. For example, two responses may receive similar quality scores while taking different interpersonal stances. In parenting advice, this matters because the answer may imply a particular balance of warmth, structure, reassurance, and boundary-setting. 

The parenting style judging was based on the Responsiveness-Demandingness framework from \citet{baumrindCurrentPatternsParental1971} and \citet{maccobySocializationContextFamily1983}. \textit{Responsiveness} captures warmth, emotional support, and sensitivity to the child’s situation. \textit{Demandingness} captures structure, expectations, behavioural guidance, and boundary-setting. Together, these dimensions define \textbf{four broad parenting styles}: authoritative, authoritarian, permissive, and uninvolved. \textit{Authoritative} advice is warm and provides structure to the child. \textit{Authoritarian} advice is firmer and less attuned. \textit{Permissive} advice is warm but loosely structured. \textit{Neglectful} advice is weak in both warmth and structure.

The style was determined through a dedicated judging prompt. The prompt defined Responsiveness and Demandingness and asked the judge to return structured JSON scores. 
Each answer receives a Responsiveness score ($R$) and a Demandingness score ($D$) from 0 to 1. These are then converted into a four-style probability distribution: authoritative = $R \times D$, authoritarian = $(1 - R) \times D$, permissive = $R \times (1 - D)$, and neglectful = $(1 - R) \times (1 - D)$, with the four probabilities normalised to sum to one. This two-step calculation is used because the four styles derive from the two dimensions. Direct classification would force mixed responses into one label, while separate $R$ and $D$ scores preserve gradation for later analysis. 

Keeping the style branch separate from rubric scoring preserves a clear distinction between advice quality and advice style, as shown in Section~\ref{sec:parentingstyle}. These scores were used to analyse how parenting style varies across models and languages.

\subsection{Automated Reporting}

After generation, quality control, and judging, the rubric outputs were exported into structured score tables and merged into a single analysis-ready matrix. The core analysis unit was a scored response indexed by scenario, model, language, and judge. 
Each row retained the generated answer, repeated judge scores, aggregated rubric means, one judge comment, and parenting style classification probabilities. This organisation supported the main comparisons used later in the paper: model comparison within a language (English vs. Chinese), language comparison within a model, and robustness comparison across judges (GPT vs. Deepseek). Summary statistics and figures visualising these comparisons were also generated, a selection of which are included in the following section.

\section{Results}
Here we present our key findings, regarding overall model performances, in addition to rubric-, scenario-, and language-specific performance comparisons (Section~\ref{sec:advicequal}), a reflection of parenting style by models (Section~\ref{sec:parentingstyle}), and judge robustness (Section~\ref{sec:judgerobustness}).

% \begin{itemize}
%     \item Advice quality
%     \begin{itemize}
%         \item Overall score
%         \begin{itemize}
%             \item Mean rubric score all models
%             \item Pairwise win-rate overall score
%         \end{itemize}
%         \item Rubric item-specific differences
%         \begin{itemize}
%             \item Radar plots for select models
%             \item Helpfulness vs safety dumbbell for all models
%         \end{itemize}
%         \item Cross-language differences
%         \begin{itemize}
%             \item Overall score differences plus table (either scatter plot or bar plot)
%             \item Rubric item-level difference heatmap
%             \item Scenario-level difference mean scores (include tags in scenario labels)
%         \end{itemize}
%     \end{itemize}
%     \item Parenting style
%     \begin{itemize}
%         \item Summary statistics and examples
%         \item Cross-language differences
%         \begin{itemize}
%             \item Style bar plots and heatmap (move heatmap to right side)
%         \end{itemize}
%         \item Rubric-linked???
%     \end{itemize}
%     \item Judge robustness
%     \begin{itemize}
%         \item Rubric judge
%         \begin{itemize}
%             \item Scatter plot of correlation of overall score
%             \item Table of correlations for rubric items?
%             \item Dumbbell plot of judge-sensitive rubrics (in English)
%         \end{itemize}
%         \item Style judge
%         \begin{itemize}
%             \item Same as above, all in English
%         \end{itemize}
%     \end{itemize}
% \end{itemize}

\subsection{Advice Quality}
\label{sec:advicequal}

%First, we present the results on the quality of LLM-generated parenting advice according to the rubric-based evaluation, beginning with overall score by model, then looking at the level of specific rubric items, before comparing advice quality across English and Chinese.

\subsubsection{4.1.1 Overall Score}

\begin{figure}[ht]
    \centering
    \includegraphics[width=\linewidth]{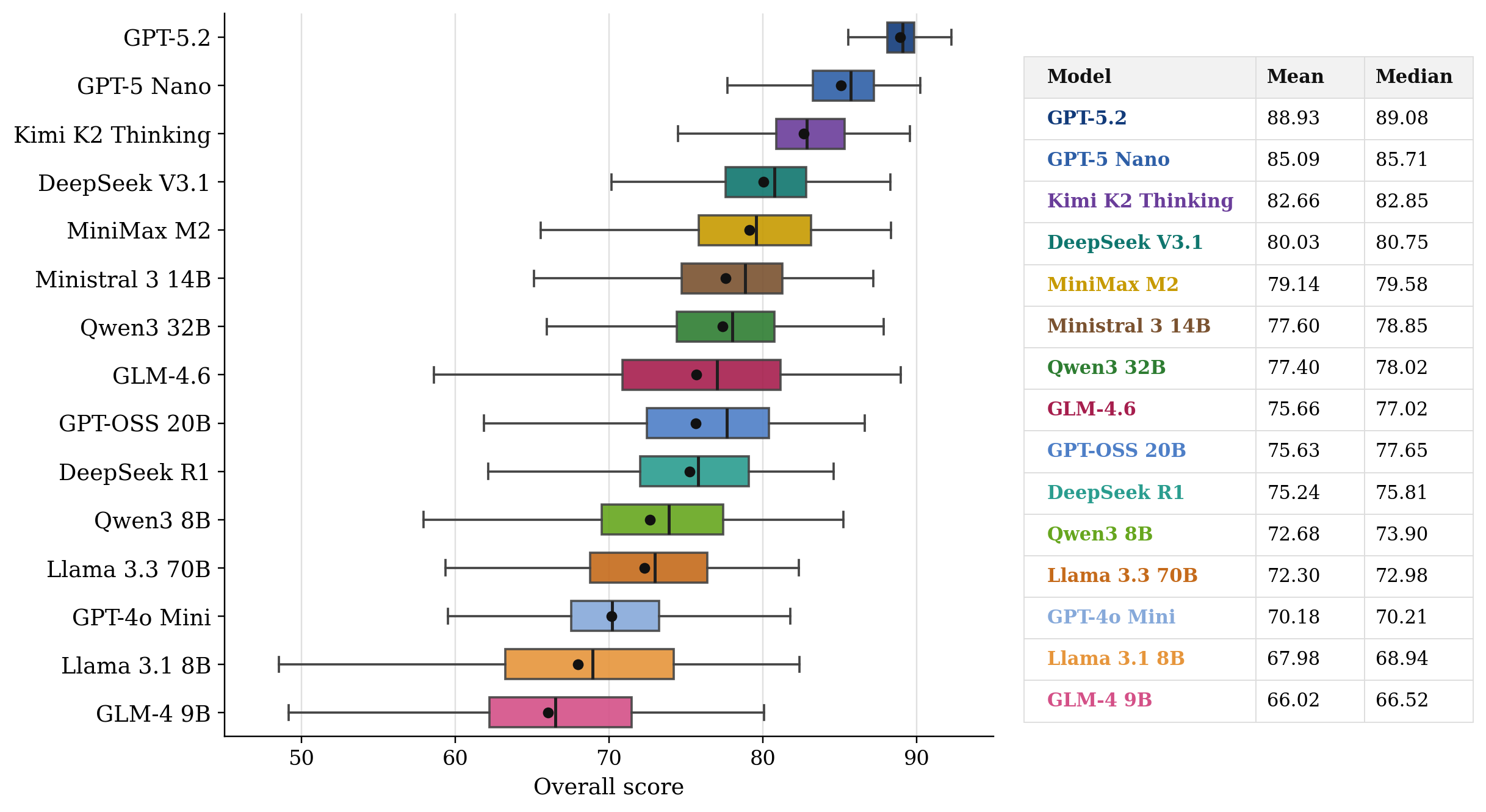}
    \caption{Overall score distribution across the scenarios by model. Primary judge = GPT-5.2}
    \label{fig:overallscore}
\end{figure}

The benchmark shows clear overall performance differences across the 15 evaluated models. Figure~\ref{fig:overallscore} reports the distribution of overall scores under the primary judge. GPT-5.2 occupies the strongest position, followed by GPT-5 Nano and Kimi K2 Thinking. GLM-4 9B and Llama 3.1 8B are lower across most of the distribution.

\begin{figure}[t]
    \centering
    \includegraphics[width=0.9\linewidth]{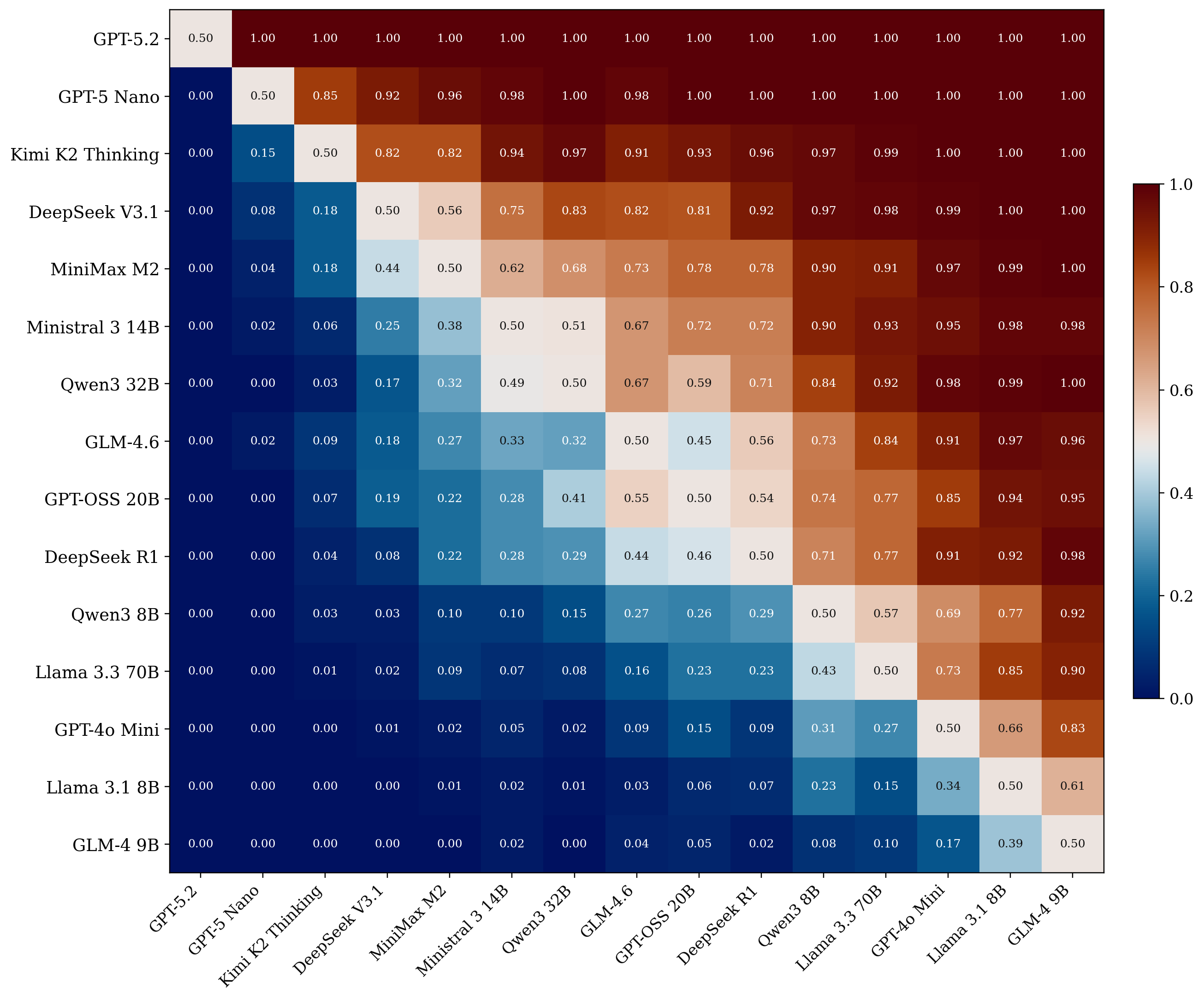}
    \caption{Heatmap of pairwise win-rate of overall rubric score for each scenario. Primary judge = GPT-5.2}
    \label{fig:overallpairwise}
\end{figure}

Pairwise win-rate tests whether model differences remain stable across scenarios. As shown in Figure~\ref{fig:overallpairwise}, GPT-5.2 dominates almost all pairwise comparisons, while GPT-5 Nano and Kimi K2 Thinking also retain strong positions. Lower-ranked models accumulate broad losses, suggesting that the benchmark captures a stable comparative structure rather than isolated favourable cases.

\subsubsection{4.1.2 Rubric-Level Scores}

Figure~\ref{fig:overallscore} establishes the broad performance tiers of the benchmark. However, overall score does not explain what kind of advice quality produces each model’s position. A model may give practical guidance while remaining weak in risk framing, or remain cautious while giving insufficient next steps. The rest of the section therefore looks inside the score profile rather than treating the leaderboard as the endpoint.

\paragraph{\textit{Model Profiles}}

\begin{figure}[]
    \centering
    \includegraphics[width=0.8\linewidth]{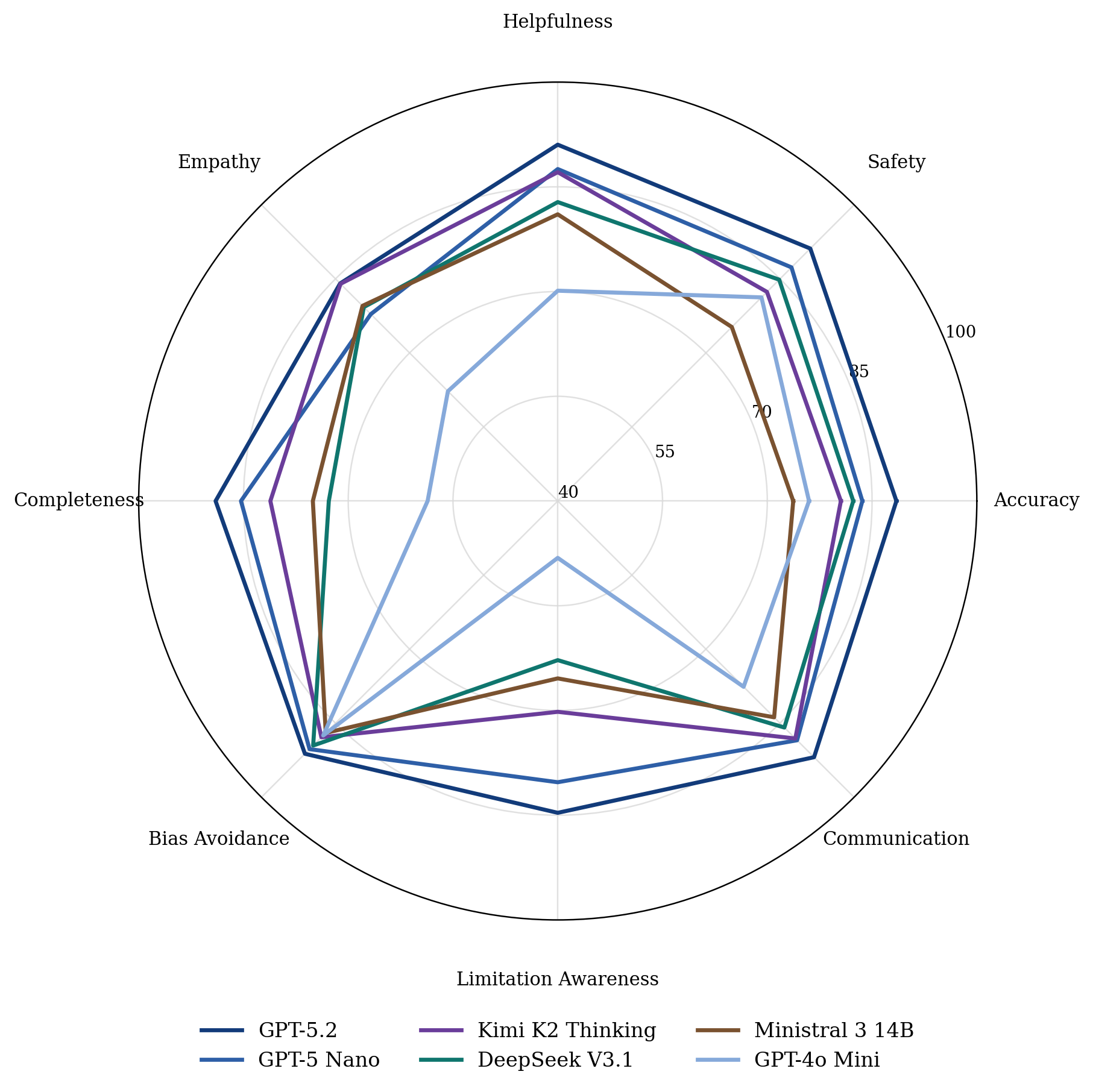}
    \caption{Radar plot showing mean rubric-item score across scenarios for selected models. Primary judge = GPT-5.2}
    \label{fig:rubricprofiles}
\end{figure}

Figure~\ref{fig:rubricprofiles} compares rubric profiles for six representative models: GPT-5.2, GPT-5 Nano, Kimi K2 Thinking, DeepSeek V3.1, Ministral 3 14B, and GPT-4o Mini. The radar plot shows how overall performance is distributed across the eight rubrics.

The main observation is that strong models do not share a single profile shape. GPT-5.2 and GPT-5 Nano are strong and balanced across rubrics. Kimi K2 Thinking is especially strong on Helpfulness, Empathy, and Communication, while DeepSeek V3.1 is relatively stronger on Accuracy, Safety, and Bias Avoidance. GPT-4o Mini shows larger deficits, especially on Completeness and Limitation Awareness.

\paragraph{\textit{Rubric-Item Comparison}}

While aggregate rubric scores provide a useful overall comparison between models, examining individual rubric dimensions reveals more specific differences in the kinds of advice models produce. Full rubric-level results are available in the appendix. We focus here on Helpfulness and Safety as representative dimensions because effective parenting advice must balance practical utility with appropriate risk management. Helpfulness captures whether a response provides clear and actionable guidance, while Safety reflects whether the response appropriately manages risk, avoids harmful suggestions, and sets suitable boundaries.

\begin{figure}
    \centering
    \includegraphics[width=\linewidth]{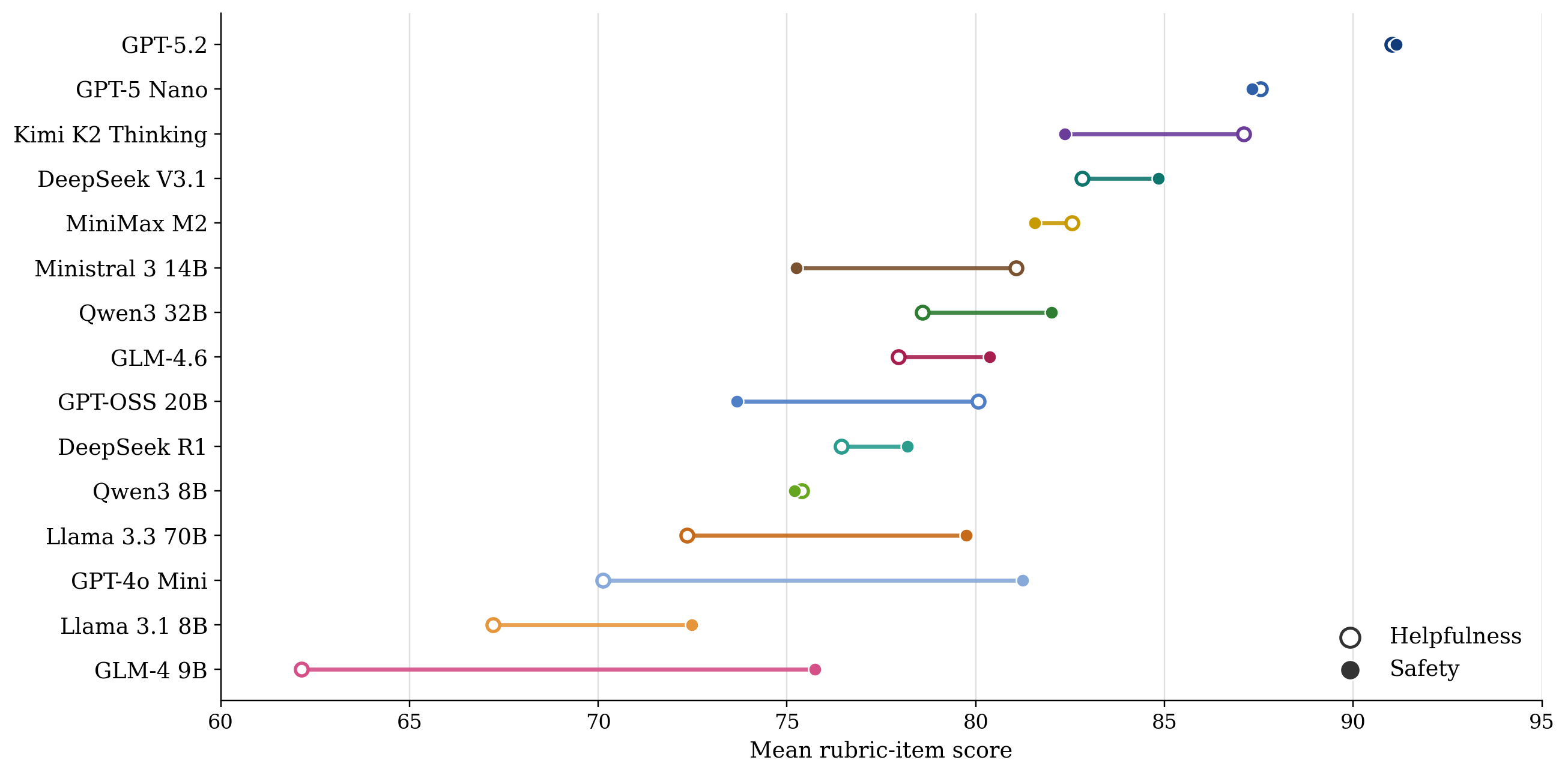}
    \caption{Dumbbell plot of mean helpfulness and safety score for each model. Primary judge = GPT-5.2}
    \label{fig:helpfulnessvssafety}
\end{figure}

Figure~\ref{fig:helpfulnessvssafety} compares each model’s mean helpfulness and safety scores. GPT-5.2 and GPT-5 Nano remain strong and relatively balanced on both dimensions. Kimi K2 Thinking also performs strongly overall, although its gap between helpfulness and safety is more visible than in the top GPT models. DeepSeek V3.1 and Qwen3 32B remain reasonably balanced, but their scores are lower than the leading models. GPT-4o Mini and Llama 3.1 8B show broader limitations, with lower scores on both dimensions.

In summary, these results show that model differences appear both in overall score and in key user-facing rubrics.

\subsubsection{4.1.3 Comparison Across Languages}

\paragraph{\textit{Overall Score Differences}}

\begin{figure}
    \centering
    \includegraphics[width=0.9\linewidth]{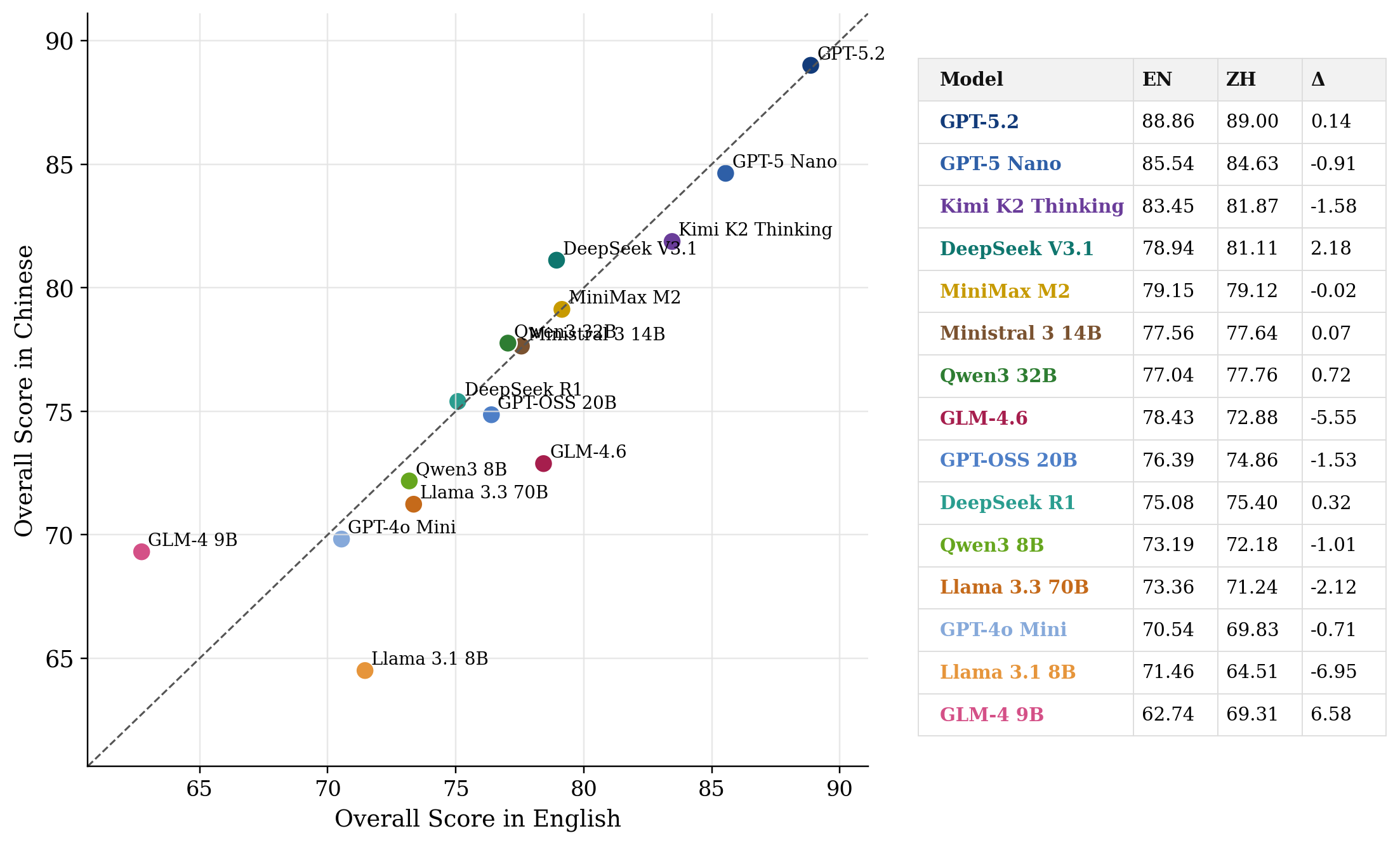}
    \caption{Scatter plot of overall score by language and model. Positive delta values and positioning above the dashed line in the scatter plot indicate higher scores in Chinese than English}
    \label{fig:crosslanguageoverall}
\end{figure}

Figure~\ref{fig:crosslanguageoverall} shows model-level cross-language differences in overall score under the primary judge. The main pattern is heterogeneity. Language-conditioned differences are present, but models separate in different directions and by different magnitudes in the English-versus-Chinese scatter.

GLM-4 9B shows the strongest positive difference in Chinese, with DeepSeek V3.1 and Qwen3 32B also moving upward. By contrast, Llama 3.1 8B and GLM-4.6 show the clearest negative differences, with Llama 3.3 70B, Kimi K2 Thinking, and GPT-OSS 20B also moving downward.

These results show a cross-language difference in advice quality that is strongly model-dependent. However, it does not yet explain which parts of the answer changed. 

\paragraph{\textit{Rubric-Level Differences}}

\begin{figure}[t]
    \centering
    \includegraphics[width=\linewidth]{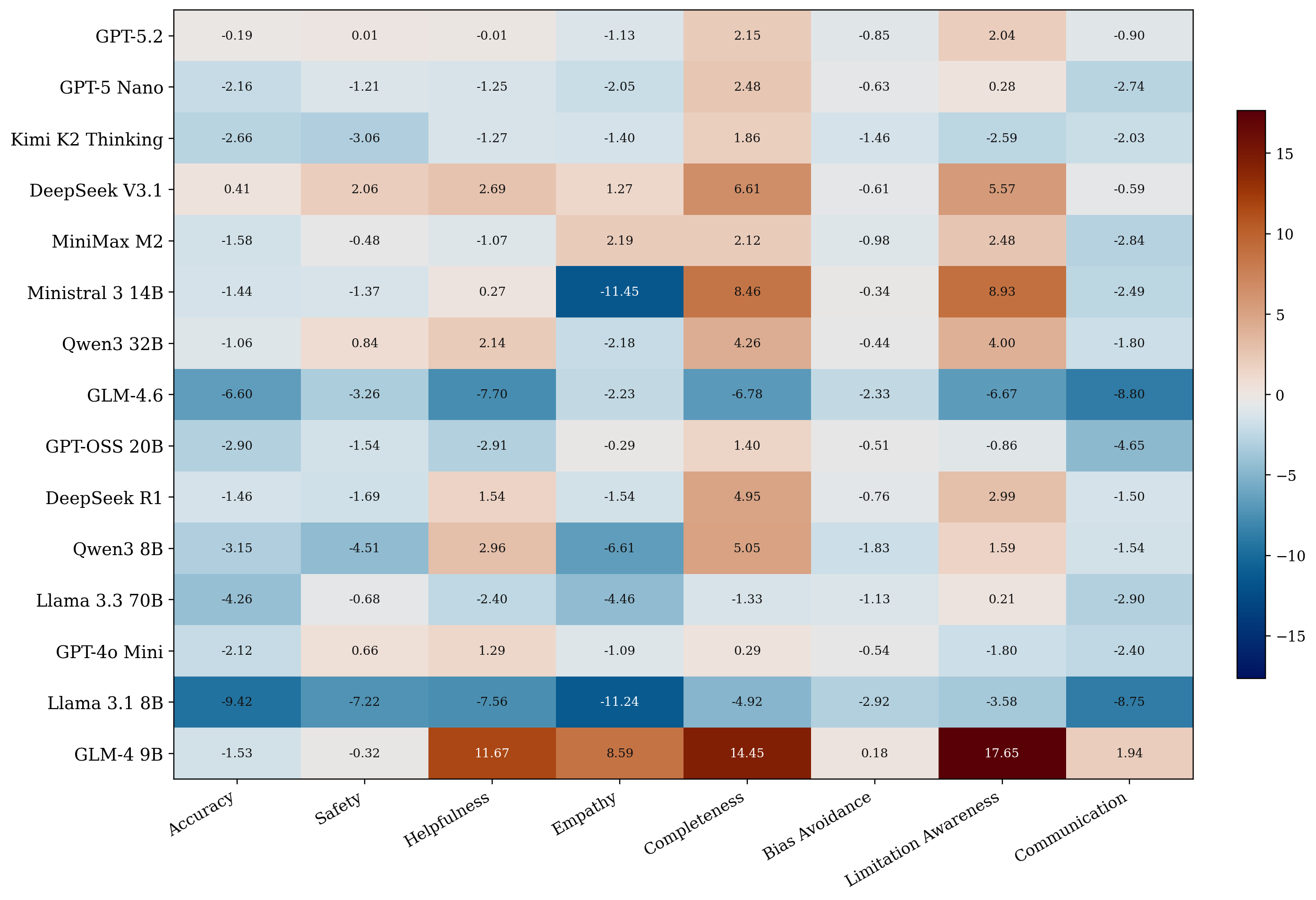}
    \caption{Heatmap of rubric-level cross-language difference by model. Primary judge = GPT-5.2. Positive values and red shading indicate higher scores in Chinese than English}
    \label{fig:crosslanguagerubric}
\end{figure}

Figure~\ref{fig:crosslanguagerubric} shows rubric-level cross-language differences under the primary judge. The clearest pattern is that Chinese answers often increase in Completeness and Limitation Awareness, while declining in Accuracy and Empathy. 

In parenting scenarios, higher Completeness suggests that Chinese responses cover more subpoints or provide more extended explanation, while higher Limitation Awareness suggests more explicit acknowledgement of uncertainty, boundaries, or professional support. Lower Empathy suggests that this expanded coverage may come with weaker reassurance or emotional attunement with the user.  

\paragraph{\textit{Scenario-Level Differences}}

\begin{figure}
    \centering
    \includegraphics[width=\linewidth]{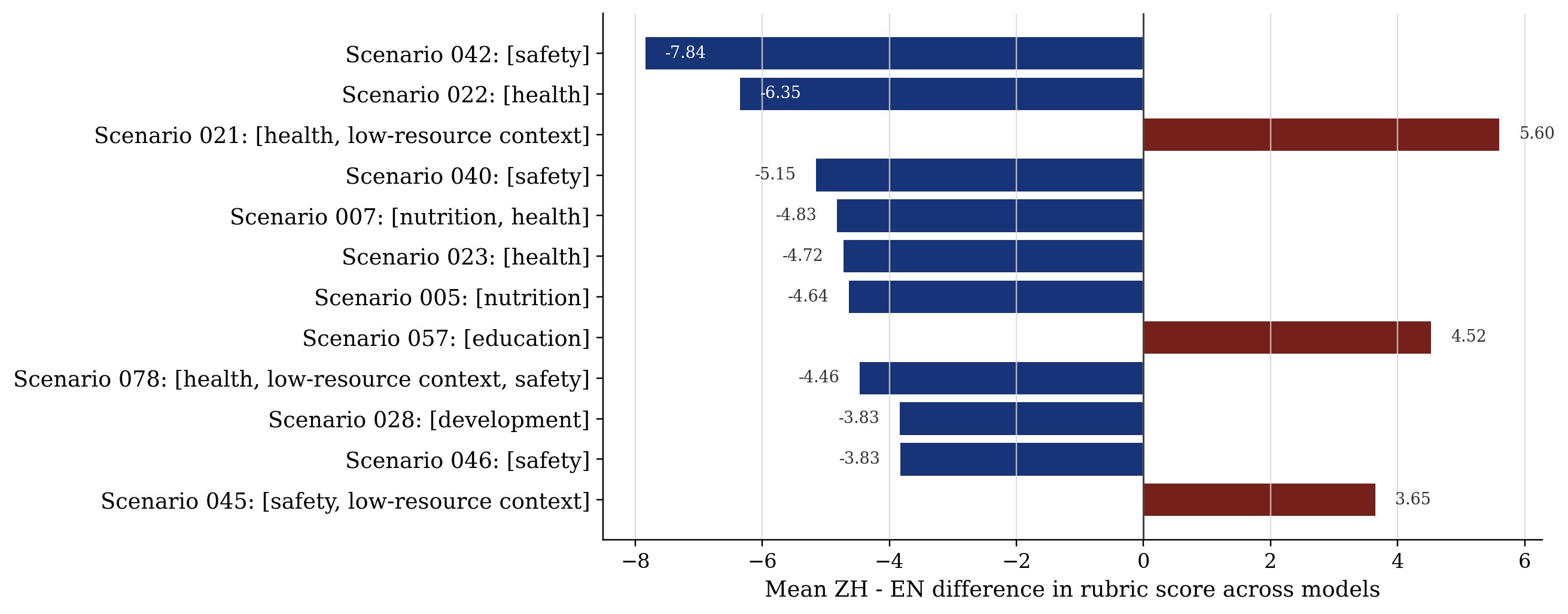}
    \caption{Bar plot for the 12 scenarios with the highest difference in overall rubric score between English and Chinese, averaged over models}
    \label{fig:crosslanguagescenario}
\end{figure}

Model-level averages show that cross-language difference exists, but they can hide where the difference occurs. Some scenarios produce little movement across most models, while others show huge language-conditioned change. Figure~\ref{fig:crosslanguagescenario} identifies the highest-difference scenarios under the primary judge.

The high-difference scenarios are concentrated in several task types. Low-resource safety and health tasks recur, including open-fire safety, river safety, emergency first aid, and possible hearing loss with limited medical access. The direction of change is not consistent; in three of these scenarios the Chinese advice is higher quality and in others the English advice is higher quality.

% The scenario-level results also distinguish directional difference from divergence. Some scenarios move many models in the same broad direction, such as Scenario 0045 and Scenario 0057. Others split models rather than producing a shared movement. Scenario 0091 is the clearest divergent case: some models improve in Chinese while others decline sharply. Language-conditioned difference is therefore both task-localised and model-dependent.

\subsection{Parenting Style}
\label{sec:parentingstyle}

\subsubsection{4.2.1 Responsiveness and Demandingness}

\begin{figure}
    \centering
    \includegraphics[width=0.9\linewidth]{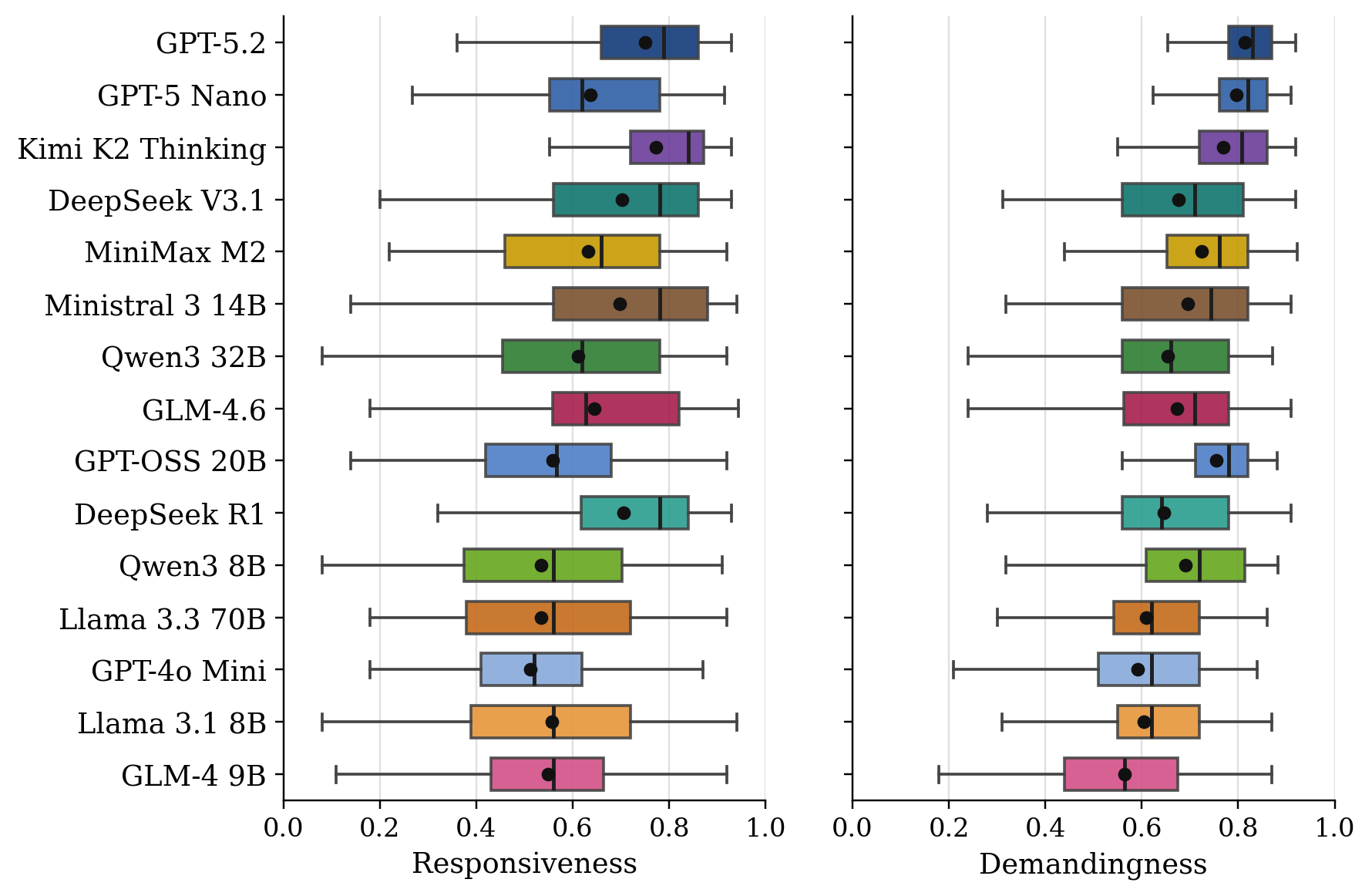}
    \caption{Overall distribution of Responsiveness and Demandingness scores across the scenarios by model. Primary judge = GPT-5.2}
    \label{fig:rd_distribution}
\end{figure}

Figure~\ref{fig:rd_distribution} reports the distribution of Responsiveness and Demandingness scores under the primary judge.
In the figure, the models are ordered top to bottom by their overall rubric score. The patterns indicate a correlation between overall score and Demandingness, i.e., high-performing models are more likely to give advice providing high levels of structure for the child compared to lower performing models. The results for Responsiveness are more mixed across models.

\subsubsection{4.2.2 Style Profiles}

\begin{figure*}[t]
    \centering
    \includegraphics[width=0.9\linewidth]{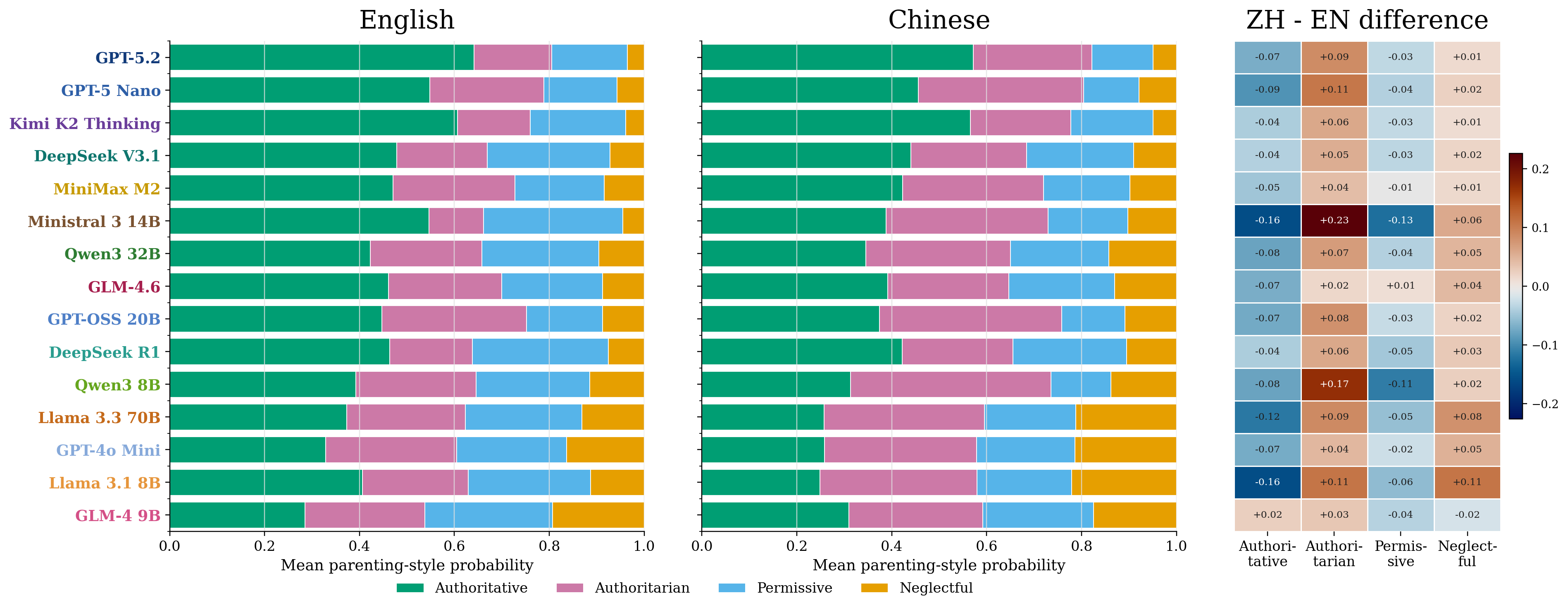}
    \caption{Stacked bar plot showing the parenting style profiles of different models across four parenting types (authoritative, authoritarian, permissive, and neglectful) and two languages (English and Chinese). Positive values and red shading in the heatmap indicate that parenting style was more prevalent in Chinese than English for the given model}
    \label{fig:parenting_style_en_zh}
\end{figure*}

Figure~\ref{fig:parenting_style_en_zh} shows the parenting-style profiles. The English style space is centred on authoritative advice. GPT-5.2 and Kimi K2 Thinking have the strongest authoritative profiles, while GPT-5 Nano carries a larger authoritarian component. Ministral 3 14B has a relatively larger permissive component among the stronger models. At the lower end, GPT-4o Mini and GLM-4 9B show weaker authoritative cores and larger neglectful shares. 
Reflecting the patterns in Demandingness across models, Figure~\ref{fig:parenting_style_en_zh} also indicates a correlation between parenting style and overall score, i.e., high-performing models are more likely to take an authoritative stance compared to lower performing models.
% \todo{Stats to justify this}

\subsubsection{4.2.3 Comparison Across Languages}

The model profiles when responding in Chinese shows a different distribution to when responding in English, indicating that language can also affect the advisory stance of the response. (Figure~\ref{fig:parenting_style_en_zh}). GPT-5.2 and Kimi K2 Thinking are still the most likely to adopt an authoritative parenting style, but both provide a greater proportion of authoritarian-style advice. The columns of the heatmap show an overall pattern in Chinese responses compared to English responses: the parenting stance adopted becomes less authoritative or permissive and more authoritarian or neglectful, suggesting advice becomes lower in warmth (Responsiveness) and somewhat higher in imposed structure (Demandingness). This is particularly the case for GPT-5 Nano, Qwen3 8B, Ministral 3 14B, and GPT-OSS 20B.

% \subsubsection{Relationship to Rubric Items}

% \todo{if got time, want to do some statistical testing relating R and D scores to rubric item scores}

\subsection{Judge Robustness}
\label{sec:judgerobustness}

\subsubsection{4.3.1 Rubric Judge}

% \begin{figure}
%     \centering
%     \includegraphics[width=0.8\linewidth]{judge_agreement_scatter.png}
%     \caption{Scatter plot of primary and secondary judge overall score for each of the 1,500 model-scenario combinations.}
%     \label{fig:judgerobustness}
% \end{figure}

\begin{figure*}
     \centering
     \hfill
     \begin{subfigure}[b]{0.28\textwidth}
         \centering
         \includegraphics[width=\textwidth]{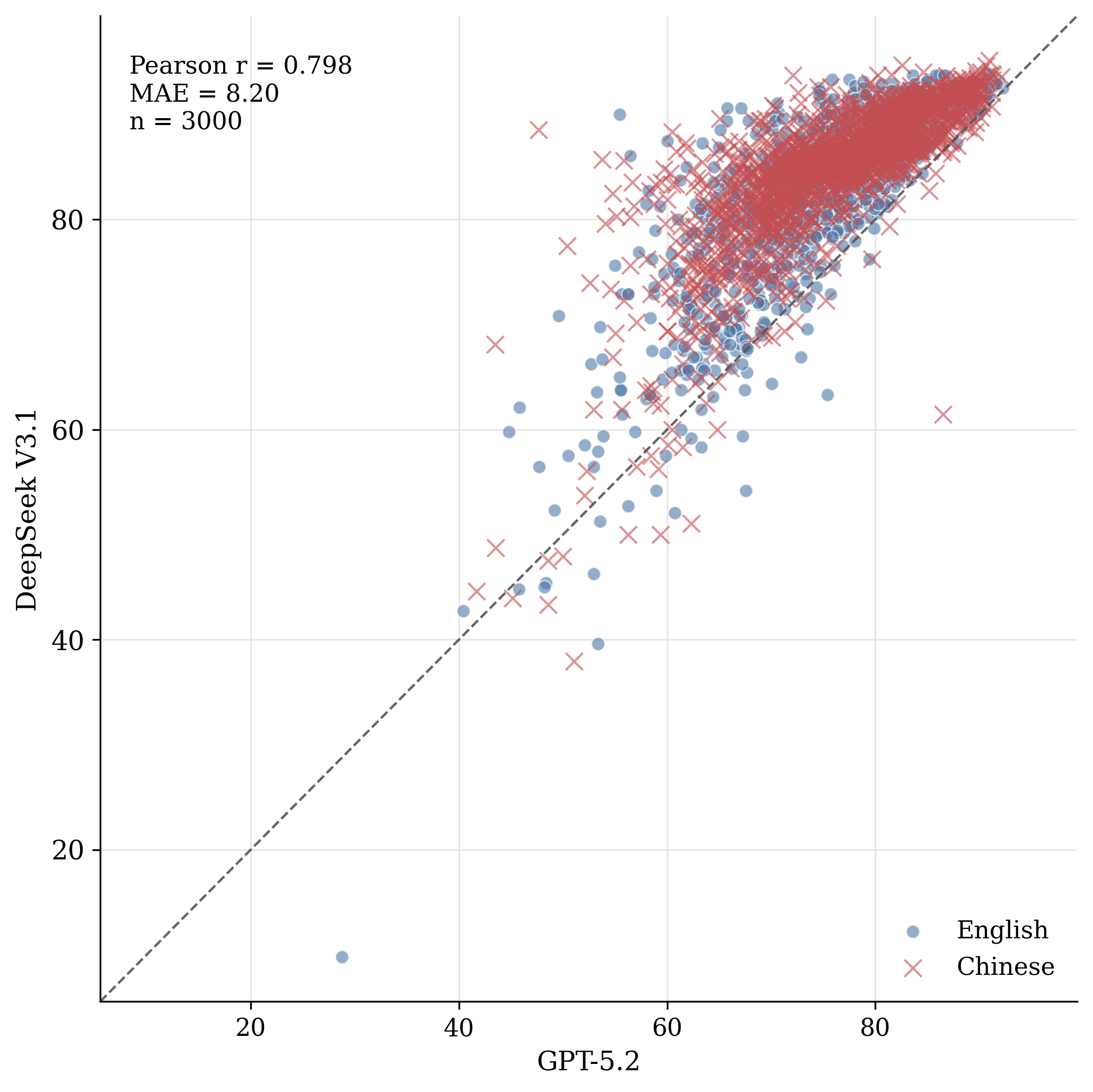}
         \caption{Overall rubric score}
         \label{fig:robustness_rubric}
     \end{subfigure}
     \hfill
     \begin{subfigure}[b]{0.28\textwidth}
         \centering
         \includegraphics[width=\textwidth]{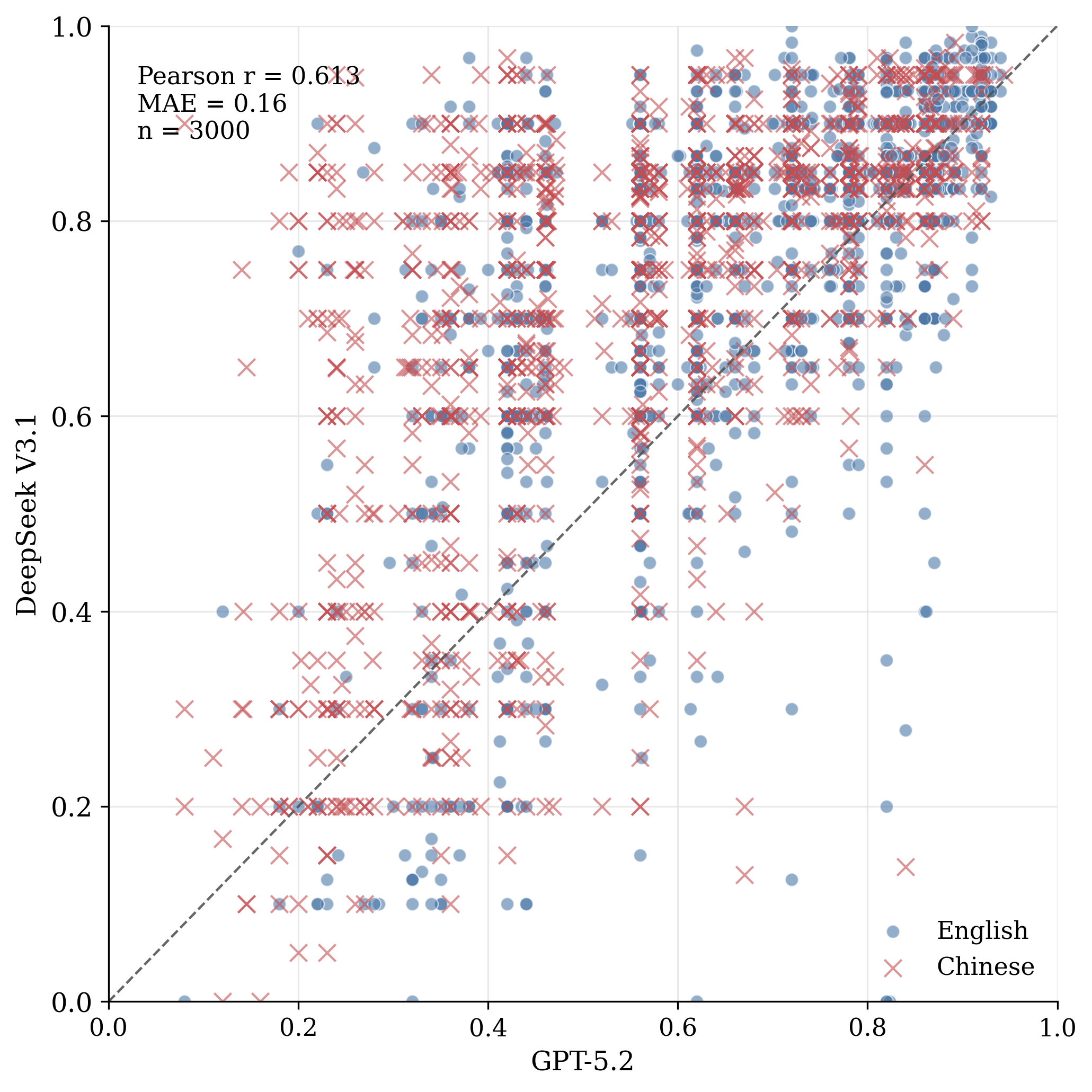}
         \caption{Responsiveness}
         \label{fig:robustness_r}
     \end{subfigure}
     \hfill
     \begin{subfigure}[b]{0.28\textwidth}
         \centering
         \includegraphics[width=\textwidth]{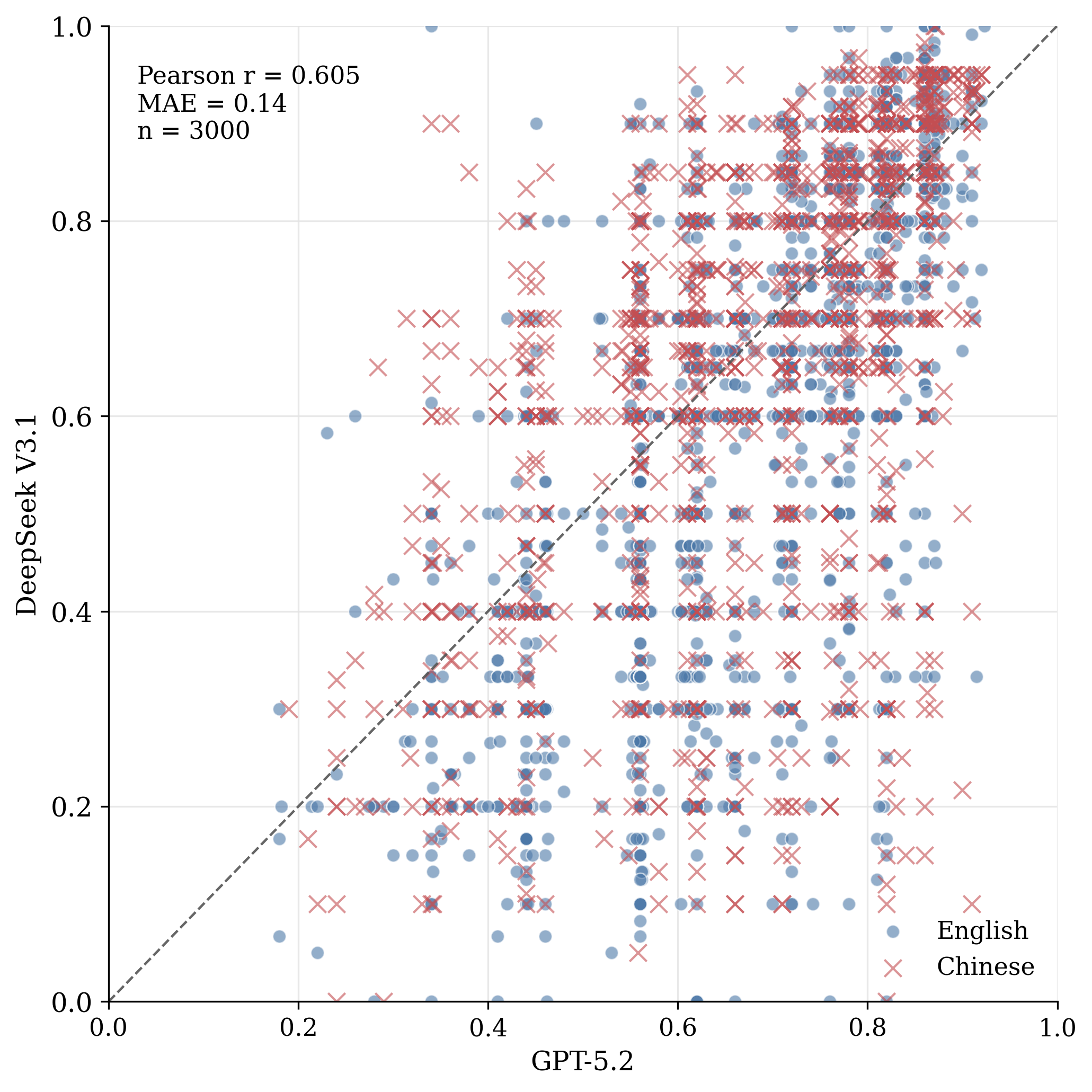}
         \caption{Demandingness}
         \label{fig:robustness_d}
     \end{subfigure}
     \hfill
        \caption{Scatter plots of primary and secondary judge scores for each of the 1,500 model-scenario combinations across English and Chinese}
        \label{fig:judgerobustness}
\end{figure*}

The main results above use GPT-5.2 as the primary judge. This section tests whether the same broad structure remains visible under the secondary judge, DeepSeek V3.1. The comparison is made on overlapping evaluation units defined by scenario, answer model, and language.
Figure~\ref{fig:robustness_rubric} visualises judge agreement on overall mean score. Across 3,000 overlapping units, the two judges reach a Pearson correlation of 0.80 and a mean absolute error of 8.2 points. This indicates a substantial shared signal, but not identical calibration. GPT-5.2 generally assigns lower overall scores than DeepSeek V3.1, with the average difference varying by answer model.

\paragraph{\textit{Judge-Sensitive Rubrics}}

Table~\ref{tab:judge_agreement_rubric} provides the comparison on the rubric-item level. Agreement between judges is strongest for Completeness, Empathy, Helpfulness, and Communication. Limitation Awareness remains relatively aligned but with a larger spread. Accuracy, Safety, and Bias Avoidance show weaker agreement and should be read as more judge-sensitive dimensions.

\begin{table}[]
    \centering
    \caption{Rubric-item level metrics comparing agreement between GPT-5.2 and DeepSeek V3.1 as judges}
    \begin{tabular}{lcc}
        \toprule
        Rubric Item & Pearson's $r$ & MAE \\
        \midrule
        Accuracy & 0.65 & 10.5 \\
        Safety & 0.61 & 8.1\\
        Helpfulness & 0.86 & 7.0\\
        Empathy & 0.87 & 12.0\\
        Completeness & 0.89& 9.1\\
        Bias Avoidance & 0.59& 3.6\\
        Limitation Awareness & 0.84& 14.3\\
        Communication & 0.81& 6.1\\
        \bottomrule
    \end{tabular}

    \label{tab:judge_agreement_rubric}
\end{table}

% \begin{figure}
%     \centering
%     \includegraphics[width=0.8\linewidth]{rd_judge_agreement__primary=gpt-5.2__secondary=deepseek-v3.1_671b-cloud__language=avg_en_zh__metric=combined.png}

%     \caption{Scatter plot of primary and secondary judge Responsiveness and Demandingness scores for each of the 1,500 model-scenario combinations. \todo{style of plot and correlations etc}}
%     \label{fig:judge_r_and_d}
% \end{figure}

\subsubsection{4.3.2 Style Judge}

Figures~\ref{fig:robustness_r} and \ref{fig:robustness_d} visualise judge agreement on Responsiveness and Demandingness scores over the 3,000 scenario-model-language combinations. For Responsiveness the judges reach a Pearson correlation of 0.61 and a mean absolute error of 0.16 points, and for Demandingness the Pearson correlation is 0.61 and mean absolute error is 0.14 points. Again, this indicates a shared signal, but not identical calibration. GPT-5.2 generally assigns higher Demandingness and lower Responsiveness scores than DeepSeek V3.1.

\section{Discussion}

\subsection{Model and Language Variations in Parenting Advice Quality and Style}

Our results show that LLM-generated parenting advice varies substantially across models, languages, and evaluation dimensions, indicating that parenting advice cannot be evaluated adequately through a single aggregate quality score. While the overall ranking identifies broad performance tiers, the rubric-level analysis shows that models reach similar total scores through different response profiles. Some models are stronger in practical guidance and communication, while others are stronger in safety, accuracy, or bias avoidance (see Section~\ref{sec:advicequal}). 
The identification of mixed profiles is important because parents may have different needs at different moments of advice-seeking: safety in high-risk situations, emotional support in times of stress, or specificity when deciding on a course of action.
Aggregate scores may oversimplify individual strength and weaknesses, potentially directly affecting user experience, particularly when the opacity of LLMs is further compounded in build user-facing applications.

Our results also suggest that supporting broader parenting advice using LLMs requires a more multi-dimensional approach. For example, a given response can be factually acceptable but poorly calibrated emotionally, or appropriately cautious but too vague to be useful. Existing evaluations of LLMs for parenting and child-related information have mainly evaluated models in health contexts, where responses can be judged against clear standards of correctness and completeness~\cite{bushuvenChatGPTCanYou2023, mcfaydenChatGPTArtificialIntelligence2024, sezginClinicalAccuracyLarge2023}, and our work extended this. Our finding aligns with arguments that LLM performance in open-ended, and thus, socially situated tasks should be assessed in relation to user-facing qualities rather than only task success or benchmark accuracy~\cite{xiaoHumanCenteredEvaluationAuditing2024}.

The cross-language results further show that advice quality is not stable across language conditions. Some models improved in Chinese, while others worsened, and these differences were not evenly distributed across rubrics or scenarios.
This highlights that multilingual evaluation should not be treated as a simple translation exercise. Even when scenario content is fixed, model behaviour may shift in ways that affect both practical advice quality and relational aspects of the response.
These patterns may reflect uneven multilingual training and alignment data~\cite{grattafioriLlama3Herd2024}, English-centric representations~\cite{wendlerLlamasWorkEnglish2024}, translation-mediated data~\cite{thompsonShockingAmountWeb2024, guoLargeLanguageModels2024}, or language-specific differences in judging~\cite{fuHowReliableMultilingual2025}.

The parenting style analysis adds a further layer here. In English, higher-performing models were more likely to produce advice classified as authoritative, which combined warmth towards the child with structure and expectations for them. In Chinese, many models shifted toward more authoritarian or neglectful profiles, suggesting lower responsiveness and stronger imposed structure. The origin of this difference is undetermined. It may reflect cultural patterns in parenting norms, linguistic differences in how advice is conventionally expressed, or model-specific training and alignment effects.
% \todo{parenting variation culturally, llm cultural sensitivity}
These results do not indicate that one language condition is better or worse, but they show that language might change the normative stance of the advice, including the implied relationship between parent and child.

Our findings have implications for the evaluation and deployment of LLMs for parenting advice. A model that performs well in English cannot be assumed to provide equivalent support in another language. Similarly, a model that scores well overall may still adopt a style of advice that does not match the values, needs, or context of a particular family. For parenting advice systems, \textbf{model selection should therefore consider not only overall quality, but also rubric-specific strengths and implicit parenting style}. Inconsistent models may be especially concerning: a model that performs well in many scenarios but fails sharply in safety- or health-related cases may create greater risk than a consistently limited model whose weaknesses are easier to anticipate.

\subsection{Parenting Advice Systems as Relational Technologies}

Parenting advice systems should be understood not only as information-retrieval or question-answering tools, but also as relational technologies~\cite{hertleinTechnologyRelationalSystems2018}. The advice from an LLM may shape how a parent interprets and responds to their child's behaviour, thereby influencing the relationship between the parent and a child.

This raises a difficult question: what counts as good parenting advice from an LLM? Advice that aligns with evidence-based child development knowledge, or advice that is culturally sensitive and responsive to the parents' values and circumstances? Both are important goals, but may be in tension. Some culturally familiar or traditional practices may be harmful, while some evidence-based recommendations may be unachievable in certain settings or communicated in a way that feels alienating or insensitive. For example, advice about discipline, boundaries, or independence may be received differently depending on cultural expectations, family structure, and beliefs about child-rearing. Therefore, a question such as ``How do I set boundaries and discipline my child?''~\cite{kimTestingCapabilityGenerative2025} cannot be evaluated as if there were a single universally accepted ideal response.

Our parenting-style analysis provides one way to make these normative dimensions visible. We do not claim that authoritative advice is always best or that any arbitrary descriptor of `parenting style' can fully capture the quality of a response. Rather, it makes explicit that LLMs do not provide value-neutral parenting guidance---there is no such thing. The models implicitly recommend particular forms of parent-child interaction, including different balances of warmth and structure.
This becomes especially relevant when considering alignment efforts, as parenting presents another example of a domain where values are plural, contested, and culturally situated.
A one-size-fits-all model of `good parenting advice' risks privileging one set of norms while presenting them as general guidance.

When it comes to designing parenting advice systems, they need to \textbf{support pluralistic values and context-sensitive interaction}. This does not mean systems should simply align with any user preference; safety remains essential, and advice should not normalise violations of children's rights. However, within safe boundaries, systems might need to adapt to different family contexts, make assumptions explicit, seeking further information where necessary, and offer a set of options rather than a single prescriptive answer.

\subsection{Robust, Auditable Evaluation Pipelines}

Through our work we highlight the importance of robust and auditable evaluation pipelines. Parenting advice is a high-stakes and socially sensitive domain, and small implementation failures can distort conclusions about model quality. In this study, response validation and repair were necessary because generation failures, truncation, wrapper text, and reasoning-like segments could otherwise be mistaken for poor advice quality. Backend metadata was useful but insufficient on its own, especially across languages---a complete Chinese answer ending with Chinese punctuation may be falsely flagged by English-style punctuation rules. This supports the need for evaluation pipelines that record not only final scores, but also the generation, cleaning, repair, and judging steps that produced them.

Auditability is also important because LLM-as-a-judge evaluation is useful but not definitive. The two judges showed substantial agreement on overall scores, suggesting that the benchmark captures a meaningful signal. However, agreement varied by rubric. Accuracy, Safety, and Bias Avoidance were more judge-sensitive than Completeness, Empathy, Helpfulness, and Communication. This pattern is important because the more judge-sensitive rubrics include dimensions that are central to harm prevention. A benchmark that reports only aggregate agreement may overstate robustness. Future evaluations should consider rubric-level judge agreements when applicable and treat judge comments as useful qualitative evidence rather than merely intermediate outputs.

\section{Limitations and Future Work}

This study has several limitations. First, although the rubric and scenarios were expert-informed, the final scoring relied on LLM judges rather than human annotators. The use of two judges provides a robustness check, but judge agreement varied by rubric, and dimensions such as safety, accuracy, and bias avoidance should be interpreted with particular caution. Future work should validate the rubric scores and parenting-style classifications against judgements from parenting experts, parents, and culturally diverse annotators, taking into account that what is regarded as a good answer may differ across families and cultural contexts.

Second, the evaluation used translated versions of the same 100 scenarios in English and Chinese. This design supports controlled cross-language comparison, but it does not fully capture parenting situations that originate within different cultural and linguistic contexts. Future work should therefore develop language-specific and culture-specific scenario sets rather than relying only on translation.

Third, the pipeline could be improved and scaled by automating remaining manual steps, including review of flagged generations, and refining rubrics and judging prompts through further iterative testing, particularly for culturally sensitive cases and high-risk scenarios where small differences in framing may have substantial implications for the advice given.

\section{Conclusion}

Overall, this work shows that evaluating LLM parenting advice requires attention to multiple forms of variation: across models, languages, rubric dimensions,parenting styles, and judges.
We developed a multilingual evaluation pipeline that combines expert-informed scenarios, response generation across 15 models with validation and repair, rubric-based LLM judging, parenting-style classification, and automated analysis, all with auditable traces. The results show that aggregate scores can obscure important differences in the kinds of advice models produce, including their relative strengths and weaknesses in safety, helpfulness, empathy, limitation awareness, and other user-facing dimensions. They also show that parenting advice is not value-neutral: models implicitly recommend different balances of warmth and structure, and these patterns shift across English and Chinese responses. More broadly, this paper highlights the need for human-centred benchmarks that capture relational and behavioural aspects of response and make normative assumptions visible. For socially sensitive advice domains, responsible evaluation requires moving beyond uni-dimensional leaderboards toward multi-dimensional and social and cultural context-aware assessment approaches.

% \section{Preparing an Anonymous Submission}

% This document details the formatting requirements for anonymous submissions. The requirements are the same as for camera ready papers but with a few notable differences:

% \begin{itemize}
%     \item Anonymous submissions must not include the author names and affiliations. Write ``Anonymous Submission'' as the ``sole author'' and leave the affiliations empty.
%     \item The PDF document's metadata should be cleared with a metadata-cleaning tool before submitting it. This is to prevent leaked information from revealing your identity.
%     \item References must be anonymized whenever the reader can infer that they are to the authors' previous work.
%     \item AAAI's copyright notice should not be included as a footer in the first page.
%     \item Only the PDF version is required at this stage. No source versions will be requested, nor any copyright transfer form.
% \end{itemize}

% You can remove the copyright notice and ensure that your names aren't shown by including \texttt{submission} option when loading the \texttt{aaai2026} package:

% \begin{quote}\begin{scriptsize}\begin{verbatim}
% \documentclass[letterpaper]{article}
% \usepackage[submission]{aaai2026}
% \end{verbatim}\end{scriptsize}\end{quote}

% The remainder of this document are the original camera-
% ready instructions. Any contradiction of the above points
% ought to be ignored while preparing anonymous submis-
% sions.

\bibliography{aaai2026, extras}

\clearpage
\onecolumn

\appendix

\section{Extended Results}

\subsection{Rubric-Level Means}
\label{app:rubric_level_scores}

\begin{table}[h]
    \centering
    \caption{Rubric-item means by model}
    \begin{tabular}{lcccccccc}
    \toprule
       Model  &  Accuracy & Safety & Helpfulness & Empathy & \makecell{Complete\\-ness} & \makecell{Bias\\Avoidance} & \makecell{Limitation\\Awareness} & \makecell{Communi\\-cation}\\
       \midrule
       GPT-5.2 & 91.6 & 91.4 & 92.5 & 87.3 & 90.0 & 90.1 & 87.4 & 93.1\\
       GPT-5 Nano & 87.9 & 88.9 & 90.2 & 82.9 & 87.4 & 89.2 & 83.7 & 89.9\\
       Kimi K2 Thinking & 86.6 & 86.7 & 90.4 & 87.9 & 84.4 & 87.8 & 77.7 & 90.7\\
       DeepSeek V3.1 & 86.2 & 87.4 & 85.6 & 84.3 & 77.2 & 89.4 & 71.3 & 88.4\\
       MiniMax M2 & 84.0 & 85.4 & 86.4 & 81.2 & 79.7 & 87.7 & 73.9 & 88.0\\
       Ministral 3 14B & 81.4 & 82.0 & 85.7 & 85.3 & 79.9 & 86.6 & 73.5 & 87.5\\
       Qwen3 32B & 82.9 & 84.8 & 82.0 & 79.8 & 74.8 & 88.4 & 70.6 & 86.2\\
       GLM-4.6 & 81.8 & 84.5 & 81.3 & 80.8 & 73.4 & 87.8 & 66.0 & 85.2\\
       GPT-OSS 20B & 80.4 & 80.4 & 85.9 & 79.3 & 79.7 & 87.0 & 70.7 & 88.0\\
       DeepSeek R1 & 80.6 & 83.0 & 80.3 & 83.5 & 73.2 & 87.1 & 67.9 & 85.0\\
       Qwen3 8B & 79.6 & 80.3 & 80.1 & 74.7 & 72.0 & 86.7 & 67.8 & 83.8\\
       Llama 3.3 70B & 79.4 & 82.0 & 75.1 & 72.2 & 67.2 & 86.9 & 64.4 & 82.1\\
       GPT-4o Mini & 79.5 & 82.8 & 72.9 & 69.7 & 63.5 & 86.6 & 56.2 & 80.9\\
       Llama 3.1 8B & 72.7 & 75.7 & 69.5 & 71.1 & 62.2 & 84.0 & 59.4 & 78.6\\
       GLM-4 9B & 73.8 & 78.3 & 64.9 & 70.3 & 57.9 & 84.0 & 53.5 & 75.8\\
    \bottomrule
    \end{tabular}
    
    \label{tab:placeholder}
\end{table}

\end{document}